\documentclass[letterpaper]{article} % DO NOT CHANGE THIS
\usepackage{aaai25}  % DO NOT CHANGE THIS
\usepackage{times}  % DO NOT CHANGE THIS
\usepackage{helvet}  % DO NOT CHANGE THIS
\usepackage{courier}  % DO NOT CHANGE THIS
\usepackage[hyphens]{url}  % DO NOT CHANGE THIS
\usepackage{graphicx} % DO NOT CHANGE THIS
\usepackage{natbib}  % DO NOT CHANGE THIS AND DO NOT ADD ANY OPTIONS TO IT
\usepackage{caption} % DO NOT CHANGE THIS AND DO NOT ADD ANY OPTIONS TO IT
\usepackage{booktabs}
\usepackage{multirow}
\usepackage{colortbl}
\usepackage{xcolor}
\usepackage{amsfonts,amssymb}
\usepackage{amsmath}
\usepackage{algorithm}
\usepackage{algorithmic}
\usepackage{hyperref}
\usepackage{color}
\usepackage{hyperref}

\hypersetup{
colorlinks=true,
linkcolor=blue,
filecolor=blue,
urlcolor=blue,
citecolor=cyan,
}
\usepackage{newfloat}
\usepackage{listings}
\DeclareCaptionStyle{ruled}{labelfont=normalfont,labelsep=colon,strut=off} % DO NOT CHANGE THIS
\floatstyle{ruled}
\newfloat{listing}{tb}{lst}{}
\floatname{listing}{Listing}
\title{Discrete Prior-Based Temporal-Coherent Content Prediction \\ for Blind Face Video Restoration}
\author{
    Lianxin Xie\textsuperscript{\rm 1}, Bingbing Zheng\textsuperscript{\rm 1}, Wen Xue\textsuperscript{\rm 1}, Yunfei Zhang\textsuperscript{\rm 1}, Le Jiang\textsuperscript{\rm 1}, Ruotao Xu\textsuperscript{\rm 2}, Si Wu\textsuperscript{\rm 1,\rm {2*}}, Hau-San Wong\textsuperscript{\rm 3*}
}
\affiliations{
    \textsuperscript{\rm 1}School of Computer Science and Engineering, South China University of Technology\\
    \textsuperscript{\rm 2}Institute of Super Robotics(Huangpu)\\
    \textsuperscript{\rm 3}Department of Computer Science, City University of Hong Kong\\
\{cslianxin.xie, 202321044369, csxuewen, cszhangyunfei, csjiangle\}@mail.scut.edu.cn\\
rtxu@superobots.com, cswusi@scut.edu.cn, cshswong@cityu.edu.hk%
    
}

\usepackage{bibentry}
\begin{document}

\maketitle

\begin{abstract}
\begin{quote}
Blind face video restoration aims to restore high-fidelity details from videos subjected to complex and unknown degradations. This task poses a significant challenge of managing temporal heterogeneity while at the same time maintaining stable face attributes. In this paper, we introduce a Discrete Prior-based Temporal-Coherent content prediction transformer to address the challenge, and our model is referred to as DP-TempCoh. Specifically, we incorporate a spatial-temporal-aware content prediction module to synthesize high-quality content from discrete visual priors, conditioned on degraded video tokens. To further enhance the temporal coherence of the predicted content, a motion statistics modulation module is designed to adjust the content, based on discrete motion priors in terms of cross-frame mean and variance. As a result, the statistics of the predicted content can match with that of real videos over time. By performing extensive experiments, we verify the effectiveness of the design elements and demonstrate the superior performance of our DP-TempCoh in both synthetically and naturally degraded video restoration. Code is available at \href{https://github.com/lxxie298/DP-TempCoh}{https://github.com/lxxie298/DP-TempCoh}.
\end{quote}
\end{abstract}

\section{Introduction}
The task of blind face video restoration(BFVR) aims to restore high-quality face image sequences from degraded videos, and the main challenge is to address various unknown degradations while at the same time preserving the dynamic consistency of the synthesized content. The degraded face videos typically exhibit temporal heterogeneity. Even if the frames of a face video are subjected to similar degradations, each restored face may appear different from the others significantly. As shown in Figure 1, state-of-the-art blind face restoration and video restoration methods, DiffBIR and FMA-Net, fail to synthesize faces with stable characteristics. 
\begin{figure}[ht]
\centering
\includegraphics[width=0.42\textwidth]{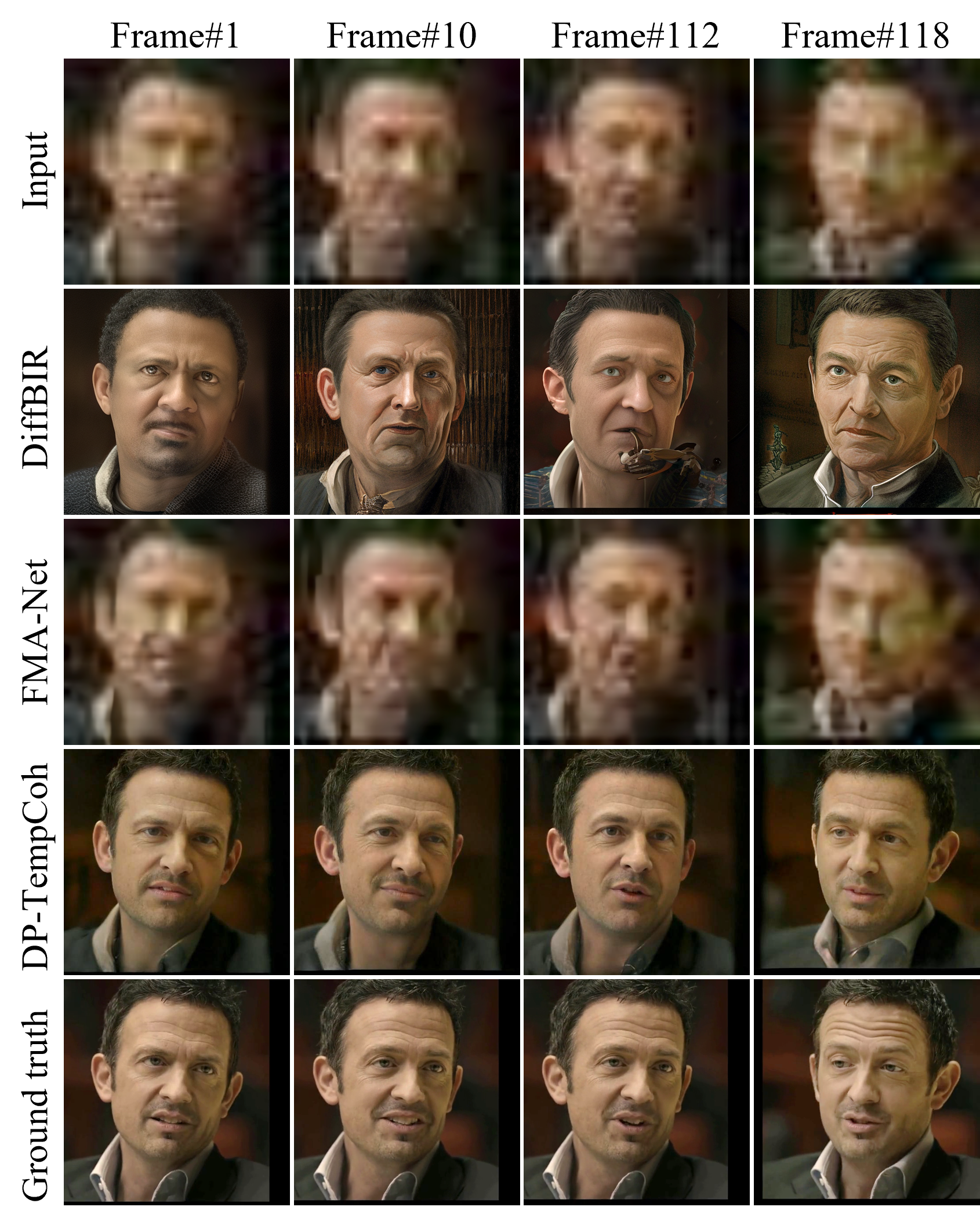}
% \vspace{-0.3cm}
\caption{An example to visually compare the proposed DP-TempCoh with the competing image/video restoration methods: DiffBIR and FMA-Net, in restoration quality and temporal coherence. }
% \vspace{-0.5cm}
\end{figure}

To synthesize clear and realistic face images from degraded ones, a number of image enhancement methods incorporate various face priors into the generation process, such as high-quality (HQ) reference samples with the same identity or geometric information\cite{li2018learning}\cite{dogan2019exemplar}. On the other hand, the well-trained Generative Adversarial Networks (GANs)\cite{chen2016infogan} on high-quality face datasets has the capability of generating diverse and realistic face images, and thus there are attempts to apply the pre-trained generator as generative priors to face image enhancement\cite{zhu2022blind}\cite{wang2021towards}\cite{wang2022panini}\cite{menon2020pulse}. On the other hand, Stable Diffusion\cite{rombach2021highresolution} has also exhibited impressive generation ability in face image generation recently \cite{lin2023diffbir}\cite{wang2023dr2}\cite{kim2022diffface}\cite{zhao2023towards}. Since the GAN- and diffusion-based methods are designed for processing images, directly applying them to face video restoration inevitably results in the synthesized results with unexpected temporal discontinuities. There are also methods that attempted to improve the temporal consistency of diffusion model, in which they fine-tuned diffusion model by adding temporal block including 3D convolution\cite{wang2024videocomposer}\cite{blattmann2023align} and temporal attention\cite{chen2023videocrafter1}\cite{zhou2022magicvideo}. Although these modifications enhance video stability, the inherent randomness of diffusion sampling can still lead to significant variations between reconstructed sequences.

In this paper, we present a Discrete Prior-based Temporal-Coherent content prediction transformer (DP-TempCoh) for high-fidelity blind face video restoration. Specifically, the proposed DP-TempCoh mainly consists of a visual prior-based spatial-temporal-aware content prediction module and a motion prior-based statistics modulation module. To to deal with the temporal heterogeneity issues in BFVR task and enhance the content consistency of the restored video, the content prediction module learns to generate the features of high-quality content from discrete visual priors in terms of latent bank, conditioned on the contextual information of degraded video frames. Further, the modulation module is incorporated to enhance the temporal coherence by modulating the predicted content, based on the motion priors in terms of cross-frame mean and variance. We have performed extensive experiments to verify the effectiveness of the designed modules and demonstrate the superior performance of our DP-TempCoh in real-world face video restoration with unknown degradation. The main contributions of this work are summarized as follows:

\begin{itemize}
  \item We propose a novel blind face video restoration framework to synthesize high-fidelity and temporal-coherent content by leveraging two types of complementary discrete priors. 

  \item We transform the spatial-temporal features of degraded video tokens by matching with the discrete visual priors in terms of a latent bank learnt from external high-quality face videos. 
  
  \item To improve the temporal coherence of synthesized face videos, we construct discrete motion priors in terms of cross-frame mean and variance, and perform modulation on the transformed video token features, such that the resulting ones can match with the temporal statistics of real face videos. 
\end{itemize}

\section{Related Work}
\subsection{Video Enhancement}
An important video enhancement task is super-resolution, which aims to restore high-resolution (HR) video frame sequences from degraded low resolution (LR) video frames while preserving the original semantics. Due to the fact that adjacent frames in video sequences can provide reference information to each other, there are a number of temporal sliding-window based VSR methods that processed LR frames in a time sliding window manner. STTN\cite{kim2018spatio}  selectively warped target frames for enhancement using the estimated optical flow. 
\cite{liu2017robust} proposed a spatial alignment network for neighboring frames alignment. Another line of research involves utilizing recurrent neural networks to harness temporal information from multiple frames. RLSP\cite{fuoli2019efficient} propagated high-dimensional hidden states to better capture long-term information. BasicVSR\cite{chan2021basicvsr} utilized a recurrent network design with bidirectional propagation to enchance video quality and detial by leveraging information from neighboring frames. BasicVSR++\cite{chan2022basicvsr++} built upon this framework by adding a second-stage refinement network and alignment module to further improve the precision and stability of the super-resolution process. Nevertheless, these methods have not incorporated suitable generative priors, which made them struggle to restore content that has lost texture details. Recently, diffusion-based methods have made great progress\cite{esser2023structure}\cite{hu2023lamd}\cite{ho2022video}. Upscale-A-Video\cite{zhou2024upscale} proposed text-guided latent diffusion framework for video enhancement, in which a flow-guided recurrent latent propagation module was used to enhance temporal stability. However, in BVFR task, it is difficult for the model to obtain stable optical flow from degraded data, which could negatively affect the performance of the restoration model.

\subsection{Blind Face Restoration} 
Significant progress has been achieved in face image restoration in recent years. There are methods that introduced facial landmarks \cite{hu2021face,lin2020learning}, face parsing maps\cite{chen2021progressive}, or 3D shapes\cite{zhu2022blind} in their designs. However, prior information obtained from degraded images often contains significant noise, which could lead to a performance drop. On the other hand, a number of recent works focus on exploring how to utilize the generative priors in generative models, such as StyleGAN2\cite{karras2020analyzing}, Stable Diffusion\cite{rombach2021highresolution}, etc, to assist in image restoration. GFP-GAN\cite{wang2021towards} and GPEN \cite{yang2021gan} leveraged facial spatial features to guide the restoration model, enabling it to utilize the geometric and texture priors of the pretrained StyleGAN. However, these methods overly relied on the facial features of degraded images, which made them sensitive to facial noise. To handle degradations robustly, DR2\cite{wang2023dr2} added Gaussian noise to the degraded images, and then restored them through a pre-trained diffusion model to achieve the effect of removing degradation. DiffBIR\cite{lin2023diffbir} first removed degradation without generating new content, and then restored face details based on a pre-trained Stable Diffusion model\cite{rombach2021highresolution}. However, because these methods were specifically designed for the task of image restoration and involve randomness in sampling, they could result in issues such as discontinuity and flickering when applied to sequences of degraded face images. There are also methods that attempted to utilize sparse representation with learned dictionaries to address image restoration tasks which took advantage of dictionary information in different ways to represent degraded images\cite{gu2022vqfr,wang2022restoreformer,zhou2022towards}. RestoreFormer learned fully-spatial interactions between corrupted queries and high-quality dictionary key-value pairs, and CodeFormer built a code sequences prediction module to exploit the dictionary information. However, these methods were specifically designed for image restoration, and did not take into account temporal coherence in the video restoration task. 

Different from the above dictionary-based methods, we model visual priors with respect to videos, capture the interrelationships among video tokens, and predict high-quality token features from the priors to replace the degraded ones. In addition, we perform cross-frame statistics modulation with motion priors, such that the predicted content can be further improved in terms of temporal coherence. This design distinguishes the proposed approach from the existing face restoration and video enhancement methods.

\section{Proposed Method}
\subsection{Overview}
We introduce a novel BFVR method of applying visual prior-based content prediction to restore degraded features, and leveraging motion prior-based statistics modulation to enhance temporal coherence. 
The structure of our framework is depicted in Figure 2 and comprises four main components: an Encoder $E$, a Visual Prior-based Spatial-temporal-aware Content Prediction Module $C$, a Motion Prior-based Statistics Modulation module $M$, and a Generator $G$. The process begins with the Encoder $E$, which transforms frames from degraded video clips $v_{lq}$ into a sequence of video tokens. These tokens $z$ are then processed by the content prediction module $C$, which generate the features $z'$ of high-quality content from a vision bank. Features $z'$ is further refined by the motion prior-based statistic modulation Module $M$, which adjusts the cross-frame mean and variance of $z'$, resulting in a motion-enhanced representation $z''$. Subsequently, several cross-attention-based transformer were applied to merge $z'$ and $z''$, forming the feature $\hat{z}$ that corresponds to a high-quality and temporally smooth face video. The feature $\hat{z}$ are then decoded by $G$ to reconstruct the face video clip $\hat{v}$ that aims to achieve maximum consistency with the ground truth. This reconstruction process contributes to the fidelity and fluidity of the restored video. 
\begin{figure*}[ht]
% \vspace{-0.3cm}
\centering
\includegraphics[width=0.95\textwidth]{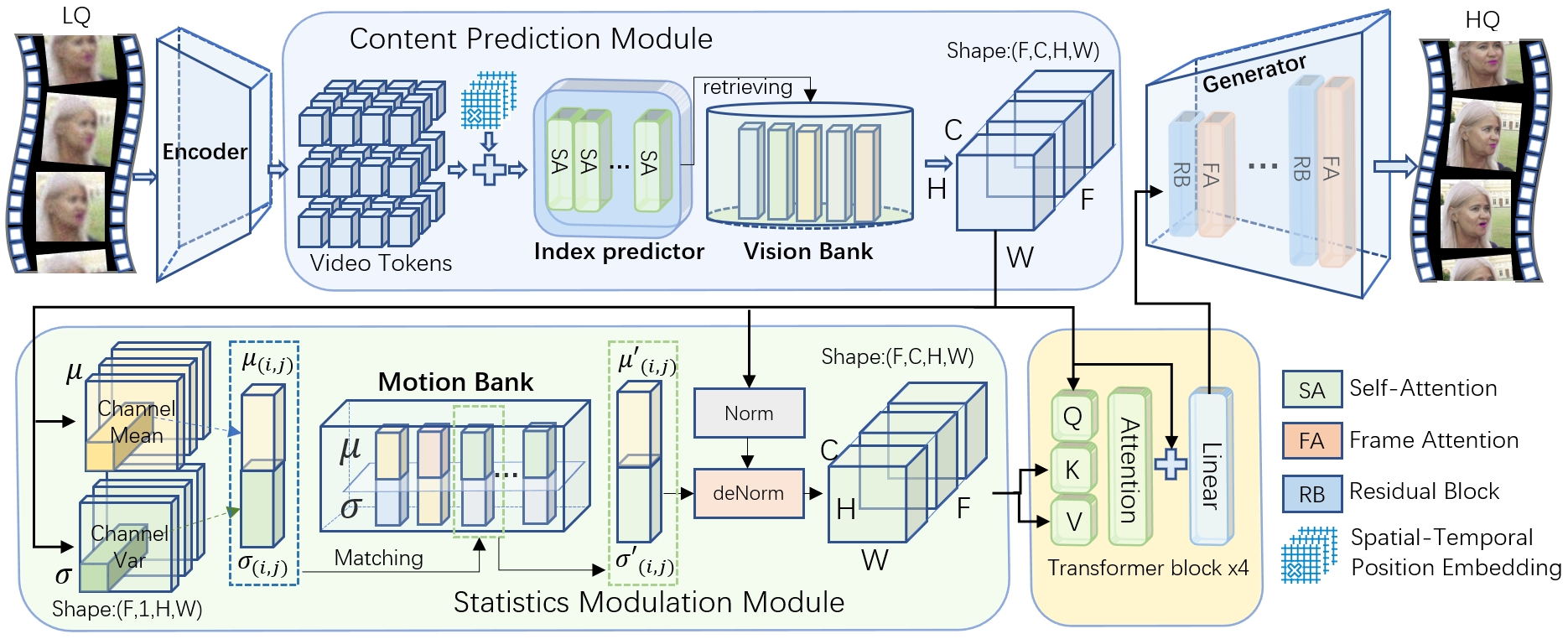}
% \vspace{-0.2cm}
\caption{Overview of the proposed DP-TempCoh framework. An encoder $E$ extracts the tokens $z$ from a degraded face video segment $v_{lq}$. A latent spatial-temporal-aware content prediction module is applied to $z$ to predict $z'$ that enriched spatial and temporal contextual information. Next, a prior-based motion statistics modulation module modulates the statistics of $z'$ to obtain $z''$. We perform several cross-attention-based transformer computation over $z'$ and $z''$, and feed the resulting feature into a generator $G$ to synthesize a HQ face video $\hat{z}$. }
\label{wild-comp}
% \vspace{-0.2cm}
\end{figure*}

\subsection{Spatial-temporal-aware Content Prediction}
Traditional image restoration techniques typically focus on individual frames, thus neglecting the dynamic inter-dependencies among successive frames, which makes it difficult to fully utilize the spatial information between frames in the video clip, especially for degraded video. 
To address this problem, the proposed spatial-temporal-aware content prediction module, conditioned on degraded video tokens, predicts index values from the pre-trained vision bank $\mathbb{T}$. The vision bank is constructed through vector quantization during the high-quality image reconstruction process, in which each index corresponds to a high-quality visual representation. In order to effectively predict index values $\hat{l}$ with nuanced contextual associations, we model both the spatial and temporal information of degraded video clips. Initially, video clip tokens $z$, extracted from encoder $E$, are flattened. To enhance the model's capability of interpreting spatial and temporal positioning of each token, we integrate a learnable spatial-temporal position embedding into each token. Subsequently, multiple self-attention-based transformer blocks, coupled with an activation function, calculate the likelihood of each token belonging to a particular bank's index. The process of index prediction can be formalized as follows:
\begin{equation}
\widetilde{z} = \xi^{'}(\texttt{SA}(\xi(z)+\mathbf{P}_{emb})), \\
\end{equation}
\begin{equation}
\hat{l}_{i,j,k} = \mathop{\arg\max}\limits_{r \in \{0,1,...,N\}} \psi(\widetilde{z}_{i,j,k})_{r},
\end{equation}
where $\mathbf{P}_{emb}$ represents the learnable spatial-temporal position embedding, $\xi$ and $\xi^{'}$ denote the flatten and unflatten operations respectively, $\texttt{SA}$ denotes the self-attention based transformer, $\psi(\cdot)_{r}$ denotes the softmax activation output on r-th index and $\hat{l}_{i,j,k} \in \{0,1,...,N\}$ denotes the predicted index of bank, where $N$ is the size of the bank $\mathbb{T}$, and $i$, $j$, and $k$ represent the spatial and temporal locations of the video clip tokens. 
% The above index prediction process ensures that the proposed module $C$ achieves a more comprehensive understanding of spatial-temporal dynamics in degraded frames. 
By computing the index $l$, we can match the corresponding features of high-quality content in the bank as follows:
\begin{equation}
z^{'}_{i,j,k} = \kappa(\mathbb{T},\hat{l}_{i,j,k}),
\end{equation} % (T,l)
where $\kappa$ represents using an index $\hat{l}_{i,j,k}$ to extract features from the bank $\mathbb{T}$ and $z^{'}$ denotes the high quality face features. The self-attention-based computation facilitates the modeling of the relationships between tokens across multiple frames, such that the module can predict index values by extracting complementary information from different tokens, eventually combining the extracted features to create a representation of high-quality content with contextual association and dynamic consistency.

\subsection{Motion Prior-based Statistics Modulation}
In order to further enhance the temporal coherence of the predicted content, we introduce the motion prior-based statistics modulation module. Initially, we construct a statistics bank, denoted as $\mathbb{M}$, which is trained using high-quality face videos. The bank $\mathbb{M}$ is designed to store the cross-frame mean and variance vectors whose components are the channel mean and variance corresponding to each frame. During the processing of features $z'$, we first calculate the mean and variance in the channel dimension for each frame as follow: 
\begin{equation}
    \mu_{f,w,h} = \sum_{c}\frac{z'_{f,c,h,w}}{L_{C}},
\end{equation}
\begin{equation}
\sigma^{2}_{f,w,h} = \frac{\sum_{c}(z'_{f,c,h,w}-\mu_{f,w,h})^{2}}{L_{C}},
\end{equation}
where $L_{C}$ denotes the channel size of $z'$. Subsequently, we concatenate $\mu_{f,w,h}$ and $\sigma^{2}_{f,w,h}$ in the temporal dimension to generate a mean vector and a variance vector, respectively. The mean and variance vectors are then concatenated, and a nearest matching algorithm is employed to retrieve the corresponding element from $\mathbb{M}$, which can be formalized as follows:
\begin{equation}
[\mu^{'}_{w,h},\sigma^{'2}_{w,h}] = \mathop{\arg\min}\limits_{e \in \mathbb{M}}||\texttt{Concat}(\mu_{w,h},\sigma^{2}_{w,h}) - e||^{2}_{2},
\end{equation}
where $\texttt{Concat}$ represents the concatenation operation in the temporal dimension. We exploit the mean $\mu^{'}_{w,h}$ and variance $\sigma^{'2}_{w,h}$ vectors with high-quality motion priors to modulate predicted content $z'$, as follows:
\begin{equation}
    z^{''}_{f,w,h} = \sigma^{'}_{f,w,h} \frac{z^{'}_{f,w,h} - \mu_{f,w,h}}{\sigma^{'}_{f,w,h} + \epsilon} + \mu^{'}_{f,w,h},
\end{equation}
where $\epsilon$ is a small real value and $z^{''}_{f,w,h}$ denotes the modulated feature. The modulation of mean and variance is tailored to maintain temporal coherence of predicted content, thereby mitigating issues like flickering. In addition, we incorporate a cross-attention based transformer block, in which feature $z'$ serves as the query and $z''$ is used as both the key and value. This arrangement allows for a comprehensive integration of content information with modulated features, which can be formalized as follows:
\begin{equation}
    \hat{z} = \xi^{'}(\texttt{CA}(\xi(z'),\xi(z''))),
\end{equation}
where $\texttt{CA}(a,b)$ denotes the transformer block in which $a,b$ denote query and key-value input, respectively. We restore the clear face video clip $\hat{v}$ by inputting $\hat{z}$ into the generator $G$. The transformer block is optimized together with G. To enhance the temporal coherence of restored texture, we have integrated multiple 3D residual blocks and frame attention mechanisms into $G$. 

\subsection{Model Training}
Let $\hat{v}$ denote the restored face video clip derived from a degraded input $v_{lq}$. Initially, we employ pixel-level loss to ensure consistency with the ground truth, defined as:
\begin{equation}
    \mathcal{L}_{consi} = \mathbb{E}_{(v,v_{hq})}[|\hat{v}-v_{hq}|_{1} + |\phi(\hat{v}) - \phi(v_{hq})|_{1} ],
\end{equation}
where $v_{hq}$ denotes the ground-truth high quality video clip with respect to $v_{lq}$, and $\phi$ represents the feature maps extracted by a pre-trained VGG19 \cite{simonyan2014very}. For the purpose of improving the visual quality of restored video segments, We also adopt an  adversarial training loss as follows:
\begin{equation}
\begin{split}
& \mathcal{L}^{real}_{adv} = \mathbb{E}_{v_{hq}}\big[\log D(v_{hq}) \big],\\
& \mathcal{L}^{sync}_{adv} = \mathbb{E}_{v_{lq}}\big[\log (1-D(\hat{v})) \big], \\
\end{split}
\end{equation}
\noindent where $D(\cdot)$ denotes the predicted probability of an input face image being real. In addition, we employ a cross entropy loss to train the spatial-temporal-aware content prediction module as follows:
\begin{equation}
    \mathcal{L}_{bank}=  \mathbb{E}_{v_{lq}}\Big[-\sum_{i,j,k}{z}_{i,j,k}^{gt}log(\psi(\widetilde{z}_{i,j,k}))\Big],
\end{equation}
where $\psi$ represents the softmax activation function and ${z}^{gt}$ denotes the ground truth of bank index labels which is derived from the pre-trained encoder $E$ and vision bank $\mathbb{T}$.
By integrating the above training losses, we formulate the optimization problem of our restoration model as follows:
\begin{equation}
\begin{split}
\min_{E, C, G} & \; \mathcal{L}_{consi} + \mathcal{L}^{sync}_{adv} + \lambda\mathcal{L}_{bank},\\
\max_{D} & \;\mathcal{L}^{real}_{adv} + \mathcal{L}^{sync}_{adv}, \\
\end{split}
\end{equation}
where $\lambda$ denotes a weighting factor that controls the relative importance of content prediction term. 
% The above training process ensures that our architecture effectively captures and leverages spatial-temporal dynamics, restoring high-quality and temporal-coherent video. 

\section{Experiments}
In this section, we conduct extensive experiments to evaluate the proposed DP-TempCoh on a wide range of blind face video restoration tasks. Initially, we outline the experimental settings, which include descriptions of the training and test datasets, implementation details, and evaluation protocol. Subsequently, we comprehensively investigate the effectiveness of the prior-based modules in face video restoration. This is followed by quantitative and qualitative comparisons with state-of-the-art methods.

\subsection{Experimental Settings}
\subsubsection{Training Data}
DP-TempCoh was trained on VFHQ\cite{xie2022vfhq}, which is a high-quality video face dataset and contains over 15,000 high-fidelity clips of diverse interview scenarios. To construct LQ training face videos, we follow\cite{yang2021gan} to degrade the VFHQ video frames as follows:
\begin{equation}
    v^{f}_{lq}=((v^{f}_{hq} \otimes \mathcal{K}_{\rho'})_{\downarrow_{b'}}+n_{\sigma'})_{JPEG_{w'}},
\end{equation}
where $v^{f}_{lq/hq}$ donotes $f$-th frame in video clip $v_{lq/hq}$. Each HQ frame is first convolved with the Gaussian blur kernel which has a standard deviation $\rho'$ $\in [\rho - 1, \rho +1]$ where $\rho$ $\in \{1 : 0.1 : 10\}$. Afterwards, it is downsampled $b' \in [b-1,b+1]$ times where $b \in \{2:32\}$, and is corrupted by Gaussian noise with intensity parameter $\sigma' \in [\sigma-1,\sigma+1]$ where $\sigma \in \{0:10\}$. Furthermore, the JPEG compression with quality factor $w' \in [w-5,w+5]$ where $w \in \{50:100\}$ is then applied to the resulting frames. 

\subsubsection{Test Data}
We assess the restoration performance of the proposed DP-TempCoh and the competing methods on three benchmark datasets: HDTF\cite{zhang2021flow}, VFHQ-Test\cite{xie2022vfhq} and YouTube Faces dataset\cite{wolf2011face}. HDTF includes about 16 hours of high-resolution videos and VFHQ-Test has 100 video clips. We randomly sample clips from each video and apply the degradation operation defined in Eq.13 to construct degraded video clips, and the resulting test dataset is referred to as HDTF-Deg and VFHQ-Test-Deg. In addition, YTF-Medium/Hard are derived from the YouTube Faces, and there are 500/520 in-the-wild face video clips with medium/heavy degradations.

\subsubsection{Implementation Details}
We implement the model using PyTorch on two NVIDIA A800s. For optimization, we use the Adam \cite{kingma2014adam} algorithm with a learning rate of \(8 \times 10^{-5}\). The training process spans 200,000 iterations with a batch size of 4. The weighting factor \(\lambda\) in Eq.12 is set to 0.5. The sizes of the vision and motion banks are 1,024 and 16,384, respectively. The size of each video clip processed by the model is 8 frames. 
% More details about the network architecture and how to pretrain the above two banks are provided in the \textbf{Appendix}. 

\subsubsection{Evaluation Protocol}
We implement all the competing methods based on the open source codes. The widely used metrics, Peak Signal-to-Noise Ratio (PSNR), Learned Perceptual Image Patch Similarity (LPIPS) and the Fréchet Inception Distances (FID) \cite{heusel2017gans}, are used to quantitatively evaluate the restored videos. 
%Considering the critical need for identity preservation in Blind Face Restoration (BFR), 
We additionally report the IDentity Similarity (IDS) based on a well-trained face recognition model: CosFace \cite{Wang2018CosFace}. Considering the temporal coherency is critical for BVFR, we further report the inter-frame difference (IFD) which evaluates video coherency by calculating the pixel mean square error between consecutive frames.

% as follows:
% \begin{equation}
%     \text{IFD} = \frac{1}{M-1}\sum_{f=2}^{M}||\hat{v}^{f}-\hat{v}^{f-1}||_{2},
% \end{equation}
% where M is the temporal length of video. A small IFD indicates that the video is more coherent. 

\subsection{Content Prediction}
In this section, we assess the effectiveness of our content prediction module. First of all, we analyze the convergence of this module. To illustrate the role of spatial-temporal context in accelerating convergence speed, we replace spatial-temporal-aware content prediction module (labeld as `S \& T-aware’) with spatial-aware version (labeled as `S-aware’) which do not consider the relationship between frames. 
\begin{figure}[ht]
% \vspace{-0.2cm}
\centering
\includegraphics[width=0.45\textwidth]{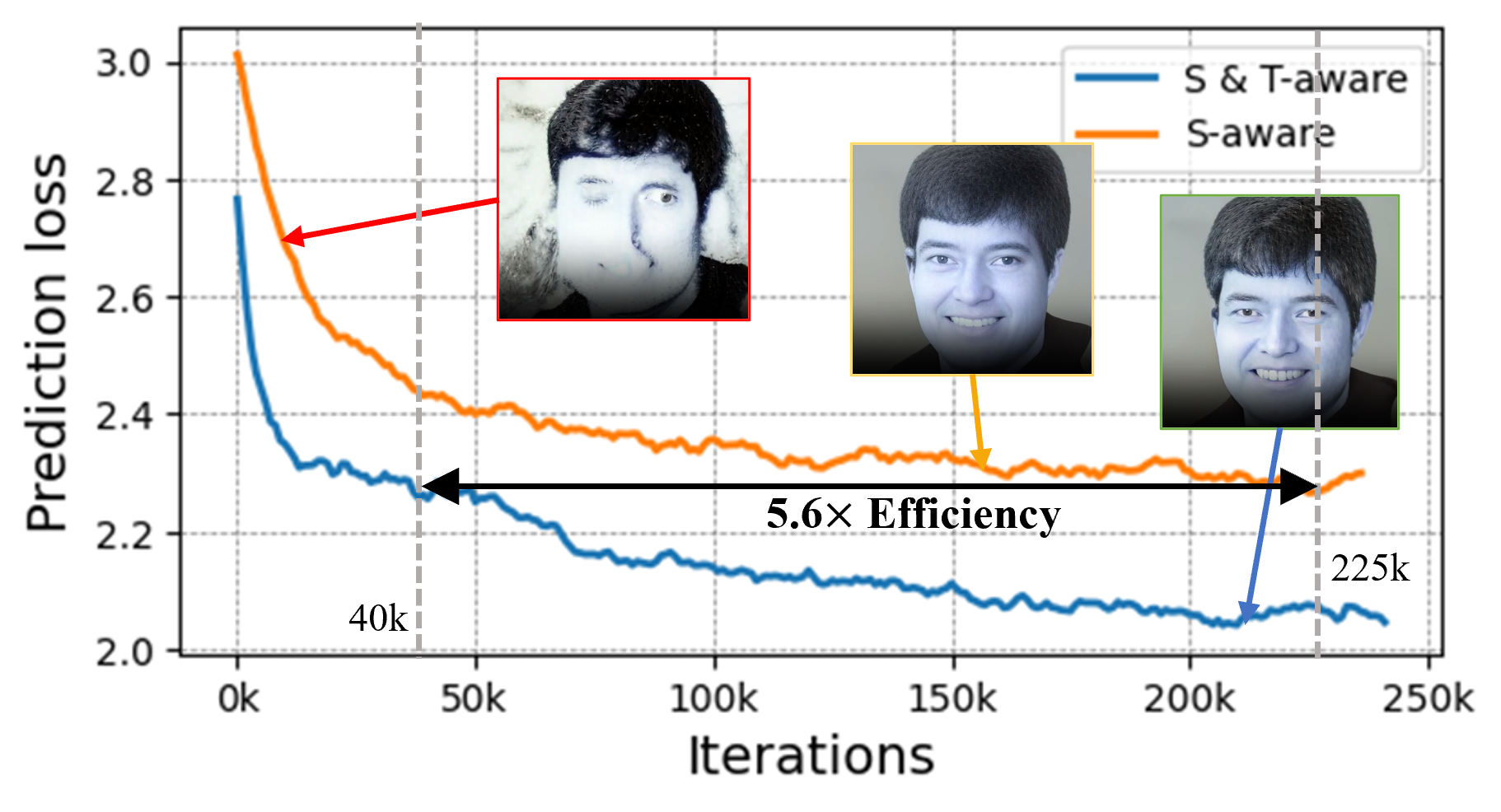}
% \vspace{-0.5cm}
\caption{Convergence comparison between spatial-temporal-aware (S \& T-aware) and spatial-aware (S-aware) prediction loss.}
\end{figure}
\begin{figure}[ht]
% \vspace{-0.4cm}
\centering
\includegraphics[width=0.45\textwidth]{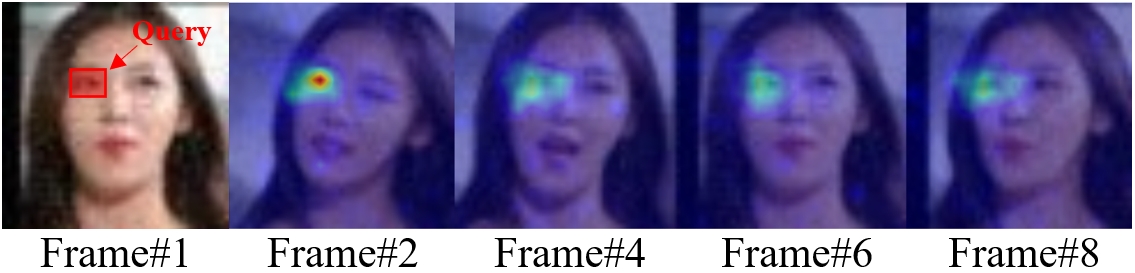}
% \vspace{-0.5cm}
\caption{Visualization of stable attention maps corresponding to the query of left eye. }
% \vspace{-0.1cm}
\end{figure}
As shown in Figure 3, at the beginning of training, distortions are noticeable in the restored face.
% As the number of iterations increased, the loss curves of the two methods began to diverge. Eventually, `S \& T-aware’ and converged to about 2.0 after 215k iterations, whereas `S-aware’ converges to 2.3 after 225k iterations. In addition, 
When converging to the same position, `S-aware’ requires 5.6 times more iterations than `S \& T-aware’. After convergence, both methods can restore face semantics well, but `S \& T-aware’ has better facial details. To ascertain whether the module leverages contextual information of degraded video frames, we visualize the attention maps of the last transformer block in Figure 4. The region of left eye serves as query, and the four attention maps on the right hand side show the responses corresponding to the query. It is worth noting that the module exhibits a heightened focus on the left eye region across different poses. The result suggests that our model effectively captures contextual information from multiple frames to restore the content on the current position. 

\subsection{Ablation Study}
%We consider that the superior restoration performance of our DP-TempCoh is mainly due to the spatial-temporal-aware content prediction and motion prior-based statistics modulation. 
% To highlight the effectiveness of the two components in BFVR, we perform a number of ablative experiments, and the results are shown in Table 1 and Figure 5.

\begin{table}[ht]
\scriptsize
\centering
    \begin{tabular}{*{8}{c}}
      \toprule
      \multirow{2}*{Exp} & \multicolumn{3}{c}{Module} & \multicolumn{3}{c}{Metrics} \\
      \cmidrule(lr){2-4}\cmidrule(lr){5-7}
      & S-aware & S \& T-aware & Motion & PSNR$\uparrow$ & FID$\downarrow$ & IFD$\downarrow$ \\
      \midrule
        (a)  & \checkmark & & & 23.12 & 72.14 & 9.86 \\
        (b)  & & \checkmark & & 24.52 & 55.11 & 3.92 \\
        (c)  & & & \checkmark & 12.55 & 210.12 & 5.59\\
        (d)  &  & \checkmark & \checkmark & \textbf{25.12} & \textbf{52.04} & \textbf{3.80} \\
      \bottomrule
    \end{tabular}
% \vspace{-0.3cm}
\caption{Results of ablative models on VFHQ-Test-Deg. ‘S-aware', ‘S \& T-aware' and ‘Motion' denote spatial-aware prediction, spatial-temporal-aware prediction and motion statistics modulation module, respectively.}
\end{table}

\begin{figure}[ht]
% \vspace{-0.3cm}
\centering
\includegraphics[width=0.45\textwidth]{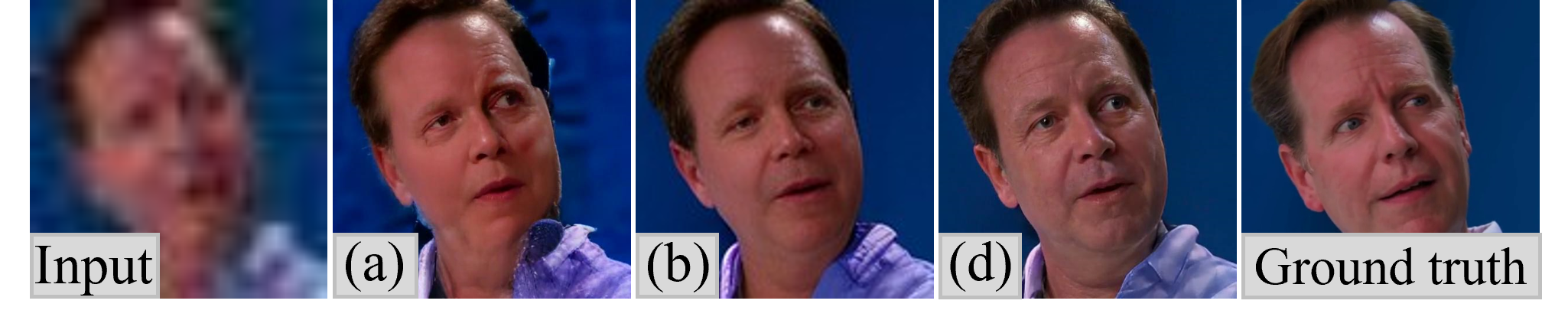}
% \vspace{-0.3cm}
\caption{Visual comparison between DP-TempCoh and ablative models(defined in Table 1) on VFHQ-Test-Deg video.}
% \vspace{-0.2cm}
\end{figure}

\subsubsection{Does prior-based spatial-temporal prediction aid restoration? }

To verify the efficacy of the proposed spatial-temporal-aware content prediction module, we conducted a comparative experiment between (a) and (b) where (a) is equipped with a spatial-aware (labeled as `S-aware’) content prediction module which does not have token information exchange between frames and (b) used the proposed spatial-temporal-aware component (labeled as `S \& T-aware’). As shown in Table 1, when using `S-aware’, all performance metrics decrease compared to the `S \& T-aware’ version, e.g., the PSNR is reduced by 1.4. This decline can be attributed to the model's inability to account for the spatial and temporal context across multiple frames. Additionally, as depicted in Figure 5, the quality of facial restoration was slightly compromised. 

% \begin{figure}[h]
% \centering
% \includegraphics[width=0.45\textwidth]{figures/bar.png}
% \vspace{-0.3cm}
% \caption{Comparison on average Euclidean distances of features between consecutive frames.}
% \vspace{-0.3cm}
% \end{figure}
\subsubsection{Can motion prior-based statistics modulation improve temporal coherence?}
We initially assess if statistics modulation enhances temporal feature coherence in latent space by comparing average Euclidean distances between consecutive frames.
\begin{table*}[ht]
% \vspace{}
\tiny
\centering
    \begin{tabular}{*{15}{c}}
      \toprule
      \multirow{2}*{Methods} & \multicolumn{5}{c}{VFHQ-Test-Deg} & \multicolumn{5}{c}{HDTF-Deg} & \multicolumn{2}{c}{YTF-Medium} & \multicolumn{2}{c}{TYF-Hard}\\
      \cmidrule(lr){2-6}\cmidrule(lr){7-11}\cmidrule(lr){12-13}\cmidrule(lr){14-15}
      & PSNR$\uparrow$ & IDS$\uparrow$ & LPIPS$\downarrow$ & FID $\downarrow$ & IFD$\downarrow$ & PSNR$\uparrow$ & IDS$\uparrow$ & LPIPS$\downarrow$ & FID $\downarrow$ & IFD$\downarrow$ & FID$\downarrow$ & IFD$\downarrow$ & FID$\downarrow$ & IFD$\downarrow$\\
      \midrule
      \rowcolor{gray!20}
      \multicolumn{15}{c}{Image Restoration Methods} \\
      \midrule
        CodeFormer \cite{zhou2022towards} & 24.75 & 0.7115 & 0.3562 & 65.68 & 9.86 & 24.55 & 0.7323 & 0.3254 & 34.14 & 8.64 & 65.15 & 7.6 & 72.25 & 9.86 \\
        RestoreFormer \cite{wang2022restoreformer} & 23.54 & 0.6034 & 0.4737 & 95.25 & 10.42 & 23.37 & 0.6543 & 0.4529 & 52.62 & 11.26 & 72.82 & 8.12 & 81.12 & 10.31 \\
        DR2 \cite{wang2023dr2} & 22.28 & 0.5411 & 0.4109 & 87.57 & 10.46 & 23.55 & 0.5916 & 0.3586 & 37.67 & 8.38 & 76.51 & 9.60 & 79.02 & 11.47 \\
        DiffBIR \cite{lin2023diffbir} & 23.96 & 0.6826 & 0.3643 & 66.85 & 11.91 & 23.09 & 0.6464 & 0.3488 & 41.25 & 12.35 & 69.74 & 10.21 & 80.18 & 13.42\\
        DifFace \cite{yue2024difface} & 25.01 & 0.6526 & 0.3483 & 63.73 & 9.82 & 24.63 & 0.6341 & 0.3281 & 34.47 & 8.75 & 68.12 & 11.10 & 83.34 & 11.48\\
        \midrule
        \rowcolor{gray!20}
        \multicolumn{15}{c}{Video Restoration Methods} \\
        \midrule
        BasicVSR++ \cite{chan2022basicvsr++} & 24.84 & 0.4802 & 0.4573 & 189.10 & 5.11 & 24.45 & 0.4851 & 0.4618 & 210.36 & 5.21 & 92.86 & 5.55 & 120.54 & 7.40 \\
        MIA-VSR \cite{zhou2024video} & 24.77 & 0.4756 & 0.5163 & 168.12 & 5.44 & 24.38 & 0.4816 & 0.5219 & 125.68 & 5.24 & 132.07 & 6.05 & 185.93 & 7.44 \\
        IA-RT \cite{xu2024enhancing} & 24.78 & 0.4754 & 0.5165 & 171.65 & 5.43 & 24.38 & 0.4814 & 0.5223 & 128.83 & 5.25 & 128.15 & 6.06 & 182.67 & 7.54\\
        FMA-Net \cite{youk2024fma} & 24.73 & 0.4464 & 0.4914 & 179.94 & 6.99 & 24.35 & 0.4515 & 0.4958 & 115.07 & 6.89 & 127.02 & 6.79 & 137.91 & 7.43\\
        \textbf{DP-TempCoh} & \textbf{25.12} & \textbf{0.7721} & \textbf{0.2262} & \textbf{52.04} & \textbf{3.80} & \textbf{24.71} & \textbf{0.7685} & \textbf{0.2427} & \textbf{25.65} & \textbf{3.13} & \textbf{51.86} & \textbf{5.51} & \textbf{55.50} & \textbf{7.38} \\
      \bottomrule
    \end{tabular}
    % \vspace{-0.3cm}
\caption{Quantitative Comparison between DP-TempCoh and competing methods on synthetic/in-the-wild data.}
\end{table*}
% As depicted in Figure 6, the degraded video tokens(labeled as `LQ’) exhibits the largest frame feature distance due to diverse degradation patterns affecting the content of each frame. After processing by the content prediction module (labeled as `+C’), there is a significant reduction in frame feature distance. The incorporation of modulation module(labeled as `+C \& M’) further improves the temporal coherence of features, which reduces the initial distance from 0.19 to 0.09, thus aligning it more closely with the frame feature distance of real face video (labeled as `GT’). 
To verify whether the modulation module can improve the temporal coherence in the video space, we replaced the S-aware module in (a) with the statistics modulation module and named it (c). As shown in Table 1, it can be seen that while the FID score increased and PSNR score decreased, IFD score decreased from 9.86 to 3.92. In experiment (d), we introduced the statistics modulation module based on experiment (b), which led to a reduction in the IFD score from 3.92 to 3.80. At the same time, both PSNR and FID metrics showed improvements. As shown in figure 5, it can be observed that both qualitative and quantitative comparisons show improvements in (d), indicating that adding the statistics modulation module to the content prediction module can further enhance the quality and temporal coherence of restored videos.

\subsection{Comparison to State-of-the-arts}
To demonstrate the superiority of the proposed DP-TempCoh, we perform quantitative and qualitative comparisons with state-of-the-arts, including the following image enhancement methods: CodeFormer\cite{zhou2022towards}, RestoreFormer\cite{wang2022restoreformer}, DR2\cite{wang2023dr2}, DiffBIR\cite{lin2023diffbir}, DifFace\cite{yue2024difface}, and video enhancement methods: BasicVSR++\cite{chan2022basicvsr++}, MIA-VSR\cite{zhou2024video}, IA-RT\cite{xu2024enhancing}, FMA-Net\cite{youk2024fma}.

\begin{figure}[ht]
\centering
\includegraphics[width=0.45\textwidth]{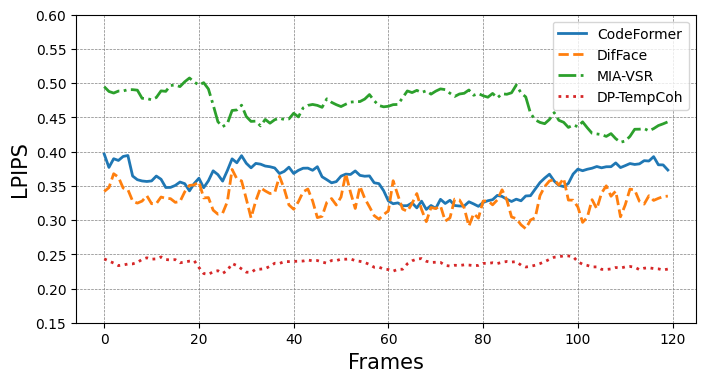}
% \vspace{-0.2cm}
\caption{Comparison of DP-TempCoh and competing methods in terms of restoration stability.}
% \vspace{-0.3cm}
\end{figure}
\begin{figure*}[ht]
% \vspace{-0.2cm}
\centering
\includegraphics[width=1.0\textwidth]{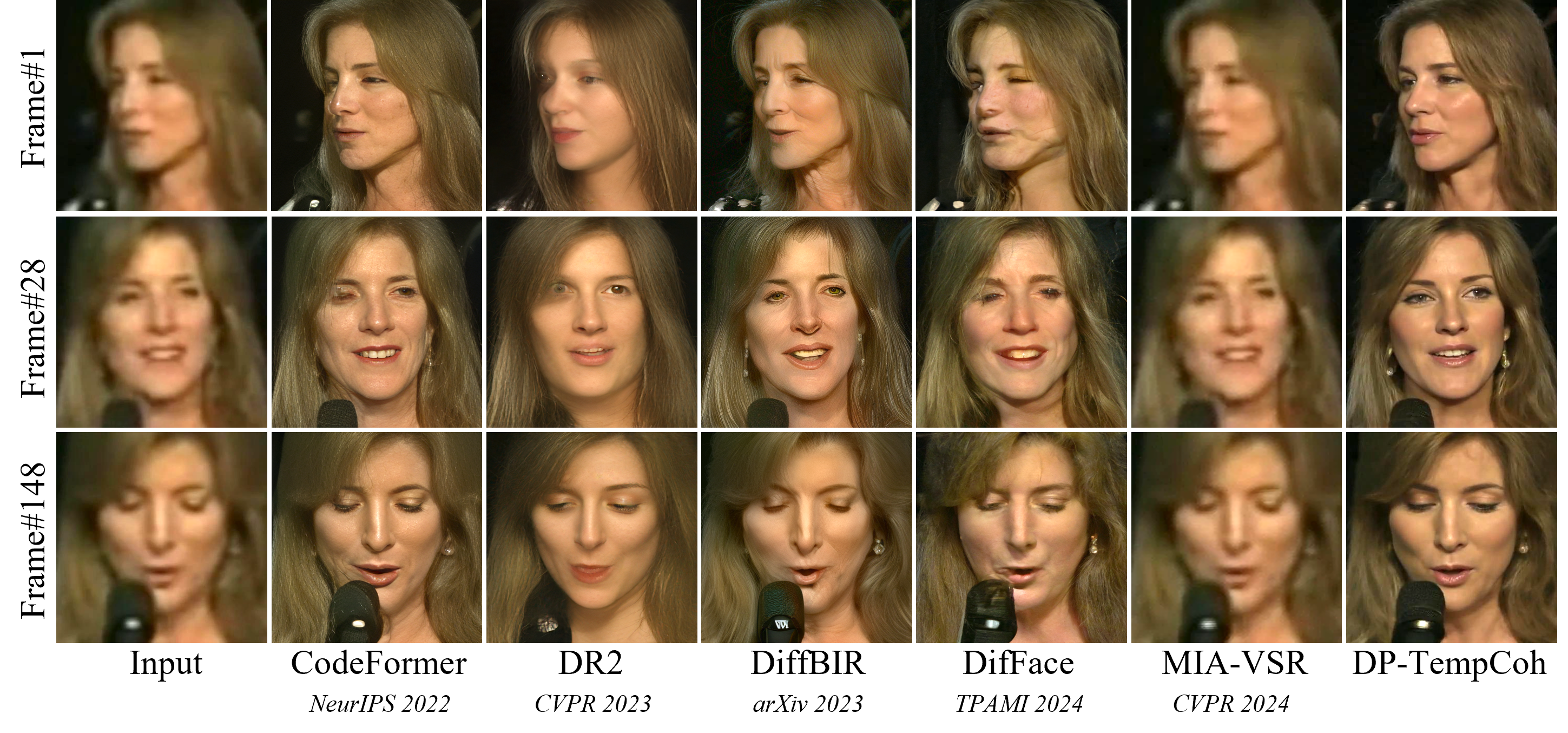}
% \vspace{-0.8cm}
\caption{Visual comparison between DP-TempCoh and the competing methods on representative frames of in-the-wild videos.}
% \vspace{-0.3cm}
\end{figure*}
\subsubsection{Results on synthetic data}
We conducted a comparative experiment in which DP-TempCoh and competing methods were used to restore degraded face videos. The results of the methods are summarized in Table 2. CodeFormer and DifFace achieve lower FID and LPIPS values than the other competing methods, which indicates that they perform better in terms of the precision and realism of the restored face sequences. In addition, the competing video restoration methods can achieve lower IFD value, suggesting that they can restore a coherent sequence of face images. On the other hand, DP-TempCoh surpasses the competing methods in terms of all the metrics. In particular, DP-TempCoh is able to achieve the highest IDS score, which reflects that it faithfully preserves the identity. In addition, DP-TempCoh achieves the IFD and FID scores of 3.13 and 25.65, which are lower than the second best methods (BasicVSR++ IDF 5.21; CodeFormer: FID 34.14) by 2.08 and 8.49, respectively. To verify whether the proposed DP-TempCoh can achieve more coherent face image sequence restoration, Figure 6 presents the LPIPS metric of the proposed method and the competing methods across a sequence of video frames. The diffusion-based method, DiffFace, shows notable fluctuations in metric LPIPS, indicating instability. Although both CodeFormer and MIA-VSR exhibit less variability, it underperform in terms of the LPIPS metric. In contrast, our proposed method consistently excels in terms of LPIPS scores and maintains stable performance across all frames. 
% The visual results on Figure 1 also demonstrates the dynamic consistency of restored frames by DP-TempCoh.
% indicate that the proposed DP-TempCoh can maintain the visual consistency of facial identity on different frames compared to DiffBIR.

% \begin{figure}[ht]
% \centering
% \includegraphics[width=0.45\textwidth]{figures/detail-cmp.png}
% \vspace{-0.3cm}
% \caption{Visual comparison between DP-TempCoh and the competing face image restoration methods on a representative in-the-wild face image.}
% \end{figure}

\begin{figure}[ht]
\centering
\includegraphics[width=0.4\textwidth]{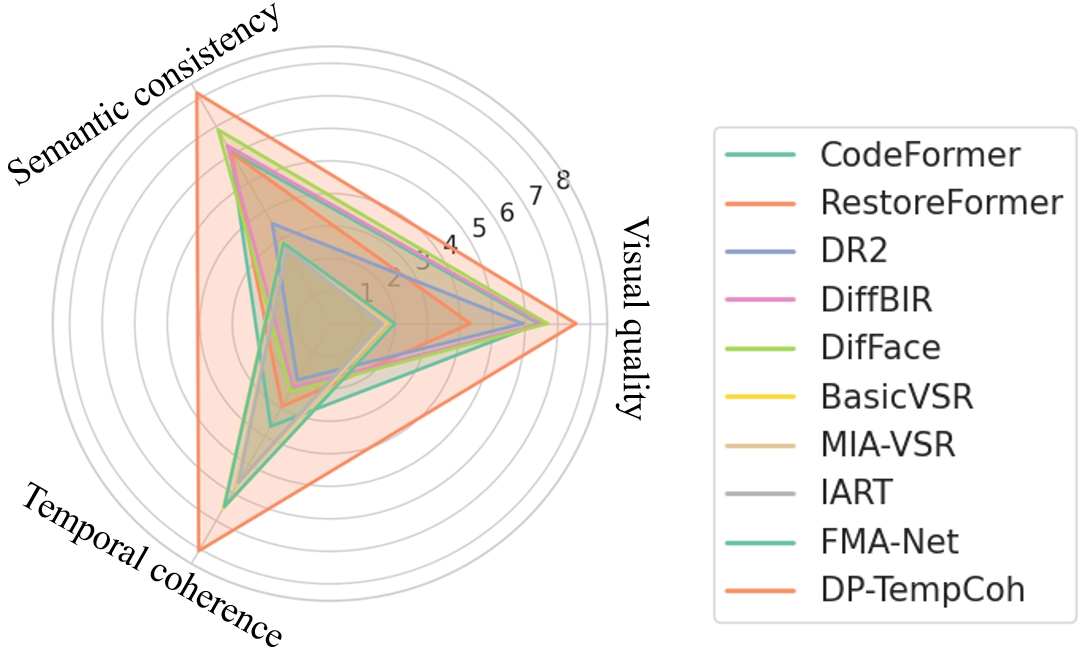}
% \vspace{-0.3cm}
\caption{The scoring result of user study on wild data.}
% \vspace{-0.5cm}
\label{userstudy}
\end{figure}

\subsubsection{Results on in-the-wild data}
To verify the generalization ability of the proposed model, we further assess DP-TempCoh and competing methods on YTF-Medium/Hard. As shown in Table 2, the DP-TempCoh model not only attains IDF metrics comparable to those of existing video restoration methods but also achieves the best FID values.
%, which is lower than the second best methods (CodeFormer 65.25) by 9.75. 
The representative results shown in Figure 7 demonstrate the advantages of DP-TempCoh in reducing artifacts, restoring realistic details, and preserving dynamic consistency. 
% We also present representative image restoration comparison results in Figure 9, in which the DP-TempCoh can restore realistic single frame face image compared to the competing image restoration methods. 
%The quantitative and qualitative comparison results demonstrate that DP-TempCoh can restore realistic and temporally coherent face videos in in-the-wild scenes compared to competing methods. 

\subsubsection{User Study}
We conduct a user study to evaluate the quality of restoration from three perspectives: visual quality, semantic consistency and tenporal coherence. We randomly select 30 degraded face video clips from YTF-Medium/Hard, and enlist 30 workers to grade the restored video from above three perspectives on a scale from 0 to 10. High scores on visual quality, semantic consistency and temporal coherence indicate realistic details, high semantic similarity with the input image and coherent video content, respectively. Figure 8 displays three average scores achieved by each model. DP-TempCoh achieves the highest average scores in terms of three perspectives.

% \subsection{Extended Applications}
% Due to the inclusion of the spatial-temporal-aware content prediction module, the proposed DP-TempCoh is capable of performing face video inpainting (FVI) based on the contextual information of the video segments. As shown in Figure 11, DP-TempCoh produces realistic and identity consistent segment. This demonstrates the greater capability of DP-TempCoh in FVI task.
% \begin{figure}[ht]
% \centering
% \includegraphics[width=0.45\textwidth]{figures/Inpainting.jpg}
% \vspace{-0.3cm}
% \caption{Representative inpainting results on HDTF video}
% \end{figure}

\section{Conclusion}
This paper presents a discrete prior-based temporal-coherent content prediction transformer to achieve high-quality face video restoration. We model the contextual information of degraded video frames to learn the interrelationships between frames, which allows us to effectively predict high-quality content from discrete visual priors. We further leverage the motion priors in terms of cross-frame mean and variance to modulate predicted content, which is essential for improving the temporal coherence of the restored video sequences. Comprehensive experimental results demonstrate that our content prediction and modulation modules can effectively generate high-quality content with dynamic consistency. A possible following up work is to apply our idea to diverse video restoration tasks, where the emergence of textual and visual prompts is concerned.

\bibliography{aaai25}

\begin{thebibliography}{44}
\providecommand{\natexlab}[1]{#1}

\bibitem[{Blattmann et~al.(2023)Blattmann, Rombach, Ling, Dockhorn, Kim,
  Fidler, and Kreis}]{blattmann2023align}
Blattmann, A.; Rombach, R.; Ling, H.; Dockhorn, T.; Kim, S.~W.; Fidler, S.; and
  Kreis, K. 2023.
\newblock Align your latents: High-resolution video synthesis with latent
  diffusion models.
\newblock In \emph{Proceedings of the IEEE/CVF Conference on Computer Vision
  and Pattern Recognition}, 22563--22575.

\bibitem[{Chan et~al.(2021)Chan, Wang, Yu, Dong, and Loy}]{chan2021basicvsr}
Chan, K.~C.; Wang, X.; Yu, K.; Dong, C.; and Loy, C.~C. 2021.
\newblock Basicvsr: The search for essential components in video
  super-resolution and beyond.
\newblock In \emph{Proceedings of the IEEE/CVF conference on computer vision
  and pattern recognition}, 4947--4956.

\bibitem[{Chan et~al.(2022)Chan, Zhou, Xu, and Loy}]{chan2022basicvsr++}
Chan, K.~C.; Zhou, S.; Xu, X.; and Loy, C.~C. 2022.
\newblock Basicvsr++: Improving video super-resolution with enhanced
  propagation and alignment.
\newblock In \emph{Proceedings of the IEEE/CVF conference on computer vision
  and pattern recognition}, 5972--5981.

\bibitem[{Chen et~al.(2021)Chen, Li, Yang, Lin, Zhang, and
  Wong}]{chen2021progressive}
Chen, C.; Li, X.; Yang, L.; Lin, X.; Zhang, L.; and Wong, K.-Y.~K. 2021.
\newblock Progressive semantic-aware style transformation for blind face
  restoration.
\newblock In \emph{Proceedings of the IEEE/CVF conference on computer vision
  and pattern recognition}, 11896--11905.

\bibitem[{Chen et~al.(2023)Chen, Xia, He, Zhang, Cun, Yang, Xing, Liu, Chen,
  Wang et~al.}]{chen2023videocrafter1}
Chen, H.; Xia, M.; He, Y.; Zhang, Y.; Cun, X.; Yang, S.; Xing, J.; Liu, Y.;
  Chen, Q.; Wang, X.; et~al. 2023.
\newblock Videocrafter1: Open diffusion models for high-quality video
  generation.
\newblock \emph{arXiv preprint arXiv:2310.19512}.

\bibitem[{Chen et~al.(2016)Chen, Duan, Houthooft, Schulman, Sutskever, and
  Abbeel}]{chen2016infogan}
Chen, X.; Duan, Y.; Houthooft, R.; Schulman, J.; Sutskever, I.; and Abbeel, P.
  2016.
\newblock {InfoGAN:} interpretable representation learning by information
  maximizing generative adversarial nets.
\newblock In \emph{Proc. Neural Information Processing Systems}.

\bibitem[{Dogan, Gu, and Timofte(2019)}]{dogan2019exemplar}
Dogan, B.; Gu, S.; and Timofte, R. 2019.
\newblock Exemplar guided face image super-resolution without facial landmarks.
\newblock In \emph{Proceedings of the IEEE/CVF conference on computer vision
  and pattern recognition workshops}, 0--0.

\bibitem[{Esser et~al.(2023)Esser, Chiu, Atighehchian, Granskog, and
  Germanidis}]{esser2023structure}
Esser, P.; Chiu, J.; Atighehchian, P.; Granskog, J.; and Germanidis, A. 2023.
\newblock Structure and content-guided video synthesis with diffusion models.
\newblock In \emph{Proceedings of the IEEE/CVF International Conference on
  Computer Vision}, 7346--7356.

\bibitem[{Fuoli, Gu, and Timofte(2019)}]{fuoli2019efficient}
Fuoli, D.; Gu, S.; and Timofte, R. 2019.
\newblock Efficient video super-resolution through recurrent latent space
  propagation.
\newblock In \emph{2019 IEEE/CVF International Conference on Computer Vision
  Workshop (ICCVW)}, 3476--3485. IEEE.

\bibitem[{Gu et~al.(2022)Gu, Wang, Xie, Dong, Li, Shan, and Cheng}]{gu2022vqfr}
Gu, Y.; Wang, X.; Xie, L.; Dong, C.; Li, G.; Shan, Y.; and Cheng, M.-M. 2022.
\newblock Vqfr: Blind face restoration with vector-quantized dictionary and
  parallel decoder.
\newblock In \emph{European Conference on Computer Vision}, 126--143. Springer.

\bibitem[{Heusel et~al.(2017)Heusel, Ramsauer, Unterthiner, Nessler, and
  Hochreiter}]{heusel2017gans}
Heusel, M.; Ramsauer, H.; Unterthiner, T.; Nessler, B.; and Hochreiter, S.
  2017.
\newblock Gans trained by a two time-scale update rule converge to a local nash
  equilibrium.
\newblock \emph{Advances in neural information processing systems}, 30.

\bibitem[{Ho et~al.(2022)Ho, Salimans, Gritsenko, Chan, Norouzi, and
  Fleet}]{ho2022video}
Ho, J.; Salimans, T.; Gritsenko, A.; Chan, W.; Norouzi, M.; and Fleet, D.~J.
  2022.
\newblock Video diffusion models.
\newblock \emph{Advances in Neural Information Processing Systems}, 35:
  8633--8646.

\bibitem[{Hu et~al.(2021)Hu, Ren, Yang, Cao, Wipf, Menze, Tong, and
  Zha}]{hu2021face}
Hu, X.; Ren, W.; Yang, J.; Cao, X.; Wipf, D.; Menze, B.; Tong, X.; and Zha, H.
  2021.
\newblock Face restoration via plug-and-play 3D facial priors.
\newblock \emph{IEEE Transactions on Pattern Analysis and Machine
  Intelligence}, 44(12): 8910--8926.

\bibitem[{Hu, Chen, and Luo(2023)}]{hu2023lamd}
Hu, Y.; Chen, Z.; and Luo, C. 2023.
\newblock Lamd: Latent motion diffusion for video generation.
\newblock \emph{arXiv preprint arXiv:2304.11603}.

\bibitem[{Karras et~al.(2020)Karras, Laine, Aittala, Hellsten, Lehtinen, and
  Aila}]{karras2020analyzing}
Karras, T.; Laine, S.; Aittala, M.; Hellsten, J.; Lehtinen, J.; and Aila, T.
  2020.
\newblock Analyzing and improving the image quality of stylegan.
\newblock In \emph{Proceedings of the IEEE/CVF conference on computer vision
  and pattern recognition}, 8110--8119.

\bibitem[{Kim et~al.(2022)Kim, Kim, Cho, Seo, Nam, Lee, Kim, and
  Lee}]{kim2022diffface}
Kim, K.; Kim, Y.; Cho, S.; Seo, J.; Nam, J.; Lee, K.; Kim, S.; and Lee, K.
  2022.
\newblock Diffface: Diffusion-based face swapping with facial guidance.
\newblock \emph{arXiv preprint arXiv:2212.13344}.

\bibitem[{Kim et~al.(2018)Kim, Sajjadi, Hirsch, and Scholkopf}]{kim2018spatio}
Kim, T.~H.; Sajjadi, M.~S.; Hirsch, M.; and Scholkopf, B. 2018.
\newblock Spatio-temporal transformer network for video restoration.
\newblock In \emph{Proceedings of the European conference on computer vision
  (ECCV)}, 106--122.

\bibitem[{Kingma and Ba(2014)}]{kingma2014adam}
Kingma, D.~P.; and Ba, J. 2014.
\newblock Adam: A method for stochastic optimization.
\newblock \emph{arXiv preprint arXiv:1412.6980}.

\bibitem[{Li et~al.(2018)Li, Liu, Ye, Zuo, Lin, and Yang}]{li2018learning}
Li, X.; Liu, M.; Ye, Y.; Zuo, W.; Lin, L.; and Yang, R. 2018.
\newblock Learning warped guidance for blind face restoration.
\newblock In \emph{Proceedings of the European conference on computer vision
  (ECCV)}, 272--289.

\bibitem[{Lin et~al.(2020)Lin, Zhang, Pan, Liu, Wang, Chen, and
  Ren}]{lin2020learning}
Lin, S.; Zhang, J.; Pan, J.; Liu, Y.; Wang, Y.; Chen, J.; and Ren, J. 2020.
\newblock Learning to deblur face images via sketch synthesis.
\newblock In \emph{Proceedings of the AAAI Conference on Artificial
  Intelligence}, 11523--11530.

\bibitem[{Lin et~al.(2023)Lin, He, Chen, Lyu, Fei, Dai, Ouyang, Qiao, and
  Dong}]{lin2023diffbir}
Lin, X.; He, J.; Chen, Z.; Lyu, Z.; Fei, B.; Dai, B.; Ouyang, W.; Qiao, Y.; and
  Dong, C. 2023.
\newblock Diffbir: Towards blind image restoration with generative diffusion
  prior.
\newblock \emph{arXiv preprint arXiv:2308.15070}.

\bibitem[{Liu et~al.(2017)Liu, Wang, Fan, Liu, Wang, Chang, and
  Huang}]{liu2017robust}
Liu, D.; Wang, Z.; Fan, Y.; Liu, X.; Wang, Z.; Chang, S.; and Huang, T. 2017.
\newblock Robust video super-resolution with learned temporal dynamics.
\newblock In \emph{Proceedings of the IEEE International Conference on Computer
  Vision}, 2507--2515.

\bibitem[{Menon et~al.(2020)Menon, Damian, Hu, Ravi, and
  Rudin}]{menon2020pulse}
Menon, S.; Damian, A.; Hu, S.; Ravi, N.; and Rudin, C. 2020.
\newblock Pulse: Self-supervised photo upsampling via latent space exploration
  of generative models.
\newblock In \emph{Proceedings of the ieee/cvf conference on computer vision
  and pattern recognition}, 2437--2445.

\bibitem[{Rombach et~al.(2021)Rombach, Blattmann, Lorenz, Esser, and
  Ommer}]{rombach2021highresolution}
Rombach, R.; Blattmann, A.; Lorenz, D.; Esser, P.; and Ommer, B. 2021.
\newblock High-Resolution Image Synthesis with Latent Diffusion Models.
\newblock arXiv:2112.10752.

\bibitem[{Simonyan and Zisserman(2014)}]{simonyan2014very}
Simonyan, K.; and Zisserman, A. 2014.
\newblock Very deep convolutional networks for large-scale image recognition.
\newblock \emph{arXiv preprint arXiv:1409.1556}.

\bibitem[{Wang et~al.(2018)Wang, Wang, Zhou, Ji, Gong, Zhou, Li, and
  Liu}]{Wang2018CosFace}
Wang, H.; Wang, Y.; Zhou, Z.; Ji, X.; Gong, D.; Zhou, J.; Li, Z.; and Liu, W.
  2018.
\newblock CosFace: Large Margin Cosine Loss for Deep Face Recognition.
\newblock In \emph{2018 IEEE/CVF Conference on Computer Vision and Pattern
  Recognition}, 5265--5274.

\bibitem[{Wang et~al.(2021)Wang, Li, Zhang, and Shan}]{wang2021towards}
Wang, X.; Li, Y.; Zhang, H.; and Shan, Y. 2021.
\newblock Towards Real-World Blind Face Restoration with Generative Facial
  Prior.
\newblock In \emph{2021 IEEE/CVF Conference on Computer Vision and Pattern
  Recognition (CVPR)}, 9164--9174.

\bibitem[{Wang et~al.(2024)Wang, Yuan, Zhang, Chen, Wang, Zhang, Shen, Zhao,
  and Zhou}]{wang2024videocomposer}
Wang, X.; Yuan, H.; Zhang, S.; Chen, D.; Wang, J.; Zhang, Y.; Shen, Y.; Zhao,
  D.; and Zhou, J. 2024.
\newblock Videocomposer: Compositional video synthesis with motion
  controllability.
\newblock \emph{Advances in Neural Information Processing Systems}, 36.

\bibitem[{Wang, Hu, and Zhang(2022)}]{wang2022panini}
Wang, Y.; Hu, Y.; and Zhang, J. 2022.
\newblock Panini-Net: GAN prior based degradation-aware feature interpolation
  for face restoration.
\newblock In \emph{Proceedings of the AAAI Conference on Artificial
  Intelligence}, 2576--2584.

\bibitem[{Wang et~al.(2022)Wang, Zhang, Chen, Wang, and
  Luo}]{wang2022restoreformer}
Wang, Z.; Zhang, J.; Chen, R.; Wang, W.; and Luo, P. 2022.
\newblock Restoreformer: High-quality blind face restoration from undegraded
  key-value pairs.
\newblock In \emph{Proceedings of the IEEE/CVF conference on computer vision
  and pattern recognition}, 17512--17521.

\bibitem[{Wang et~al.(2023)Wang, Zhang, Zhang, Zheng, Zhou, Zhang, and
  Wang}]{wang2023dr2}
Wang, Z.; Zhang, Z.; Zhang, X.; Zheng, H.; Zhou, M.; Zhang, Y.; and Wang, Y.
  2023.
\newblock Dr2: Diffusion-based robust degradation remover for blind face
  restoration.
\newblock In \emph{Proceedings of the IEEE/CVF Conference on Computer Vision
  and Pattern Recognition}, 1704--1713.

\bibitem[{Wolf, Hassner, and Maoz(2011)}]{wolf2011face}
Wolf, L.; Hassner, T.; and Maoz, I. 2011.
\newblock Face recognition in unconstrained videos with matched background
  similarity.
\newblock In \emph{CVPR 2011}, 529--534. IEEE.

\bibitem[{Xie et~al.(2022)Xie, Wang, Zhang, Dong, and Shan}]{xie2022vfhq}
Xie, L.; Wang, X.; Zhang, H.; Dong, C.; and Shan, Y. 2022.
\newblock Vfhq: A high-quality dataset and benchmark for video face
  super-resolution.
\newblock In \emph{Proceedings of the IEEE/CVF Conference on Computer Vision
  and Pattern Recognition}, 657--666.

\bibitem[{Xu et~al.(2024)Xu, Yu, Wang, Mi, and Yao}]{xu2024enhancing}
Xu, K.; Yu, Z.; Wang, X.; Mi, M.~B.; and Yao, A. 2024.
\newblock Enhancing Video Super-Resolution via Implicit Resampling-based
  Alignment.
\newblock In \emph{Proceedings of the IEEE/CVF Conference on Computer Vision
  and Pattern Recognition}, 2546--2555.

\bibitem[{Yang et~al.(2021)Yang, Ren, Xie, and Zhang}]{yang2021gan}
Yang, T.; Ren, P.; Xie, X.; and Zhang, L. 2021.
\newblock Gan prior embedded network for blind face restoration in the wild.
\newblock In \emph{Proceedings of the IEEE/CVF conference on computer vision
  and pattern recognition}, 672--681.

\bibitem[{Youk, Oh, and Kim(2024)}]{youk2024fma}
Youk, G.; Oh, J.; and Kim, M. 2024.
\newblock FMA-Net: Flow-Guided Dynamic Filtering and Iterative Feature
  Refinement with Multi-Attention for Joint Video Super-Resolution and
  Deblurring.
\newblock In \emph{Proceedings of the IEEE/CVF Conference on Computer Vision
  and Pattern Recognition}, 44--55.

\bibitem[{Yue and Loy(2024)}]{yue2024difface}
Yue, Z.; and Loy, C.~C. 2024.
\newblock Difface: Blind face restoration with diffused error contraction.
\newblock \emph{IEEE Transactions on Pattern Analysis and Machine
  Intelligence}.

\bibitem[{Zhang et~al.(2021)Zhang, Li, Ding, and Fan}]{zhang2021flow}
Zhang, Z.; Li, L.; Ding, Y.; and Fan, C. 2021.
\newblock Flow-guided one-shot talking face generation with a high-resolution
  audio-visual dataset.
\newblock In \emph{Proceedings of the IEEE/CVF Conference on Computer Vision
  and Pattern Recognition}, 3661--3670.

\bibitem[{Zhao et~al.(2023)Zhao, Hou, Su, Jia, Li, and
  Grundmann}]{zhao2023towards}
Zhao, Y.; Hou, T.; Su, Y.-C.; Jia, X.; Li, Y.; and Grundmann, M. 2023.
\newblock Towards authentic face restoration with iterative diffusion models
  and beyond.
\newblock In \emph{Proceedings of the IEEE/CVF International Conference on
  Computer Vision}, 7312--7322.

\bibitem[{Zhou et~al.(2022{\natexlab{a}})Zhou, Wang, Yan, Lv, Zhu, and
  Feng}]{zhou2022magicvideo}
Zhou, D.; Wang, W.; Yan, H.; Lv, W.; Zhu, Y.; and Feng, J. 2022{\natexlab{a}}.
\newblock Magicvideo: Efficient video generation with latent diffusion models.
\newblock \emph{arXiv preprint arXiv:2211.11018}.

\bibitem[{Zhou et~al.(2022{\natexlab{b}})Zhou, Chan, Li, and
  Loy}]{zhou2022towards}
Zhou, S.; Chan, K.; Li, C.; and Loy, C.~C. 2022{\natexlab{b}}.
\newblock Towards robust blind face restoration with codebook lookup
  transformer.
\newblock \emph{Advances in Neural Information Processing Systems}, 35:
  30599--30611.

\bibitem[{Zhou et~al.(2024{\natexlab{a}})Zhou, Yang, Wang, Luo, and
  Loy}]{zhou2024upscale}
Zhou, S.; Yang, P.; Wang, J.; Luo, Y.; and Loy, C.~C. 2024{\natexlab{a}}.
\newblock Upscale-A-Video: Temporal-Consistent Diffusion Model for Real-World
  Video Super-Resolution.
\newblock In \emph{Proceedings of the IEEE/CVF Conference on Computer Vision
  and Pattern Recognition}, 2535--2545.

\bibitem[{Zhou et~al.(2024{\natexlab{b}})Zhou, Zhang, Zhao, Wang, Li, and
  Gu}]{zhou2024video}
Zhou, X.; Zhang, L.; Zhao, X.; Wang, K.; Li, L.; and Gu, S. 2024{\natexlab{b}}.
\newblock Video Super-Resolution Transformer with Masked Inter\&Intra-Frame
  Attention.
\newblock In \emph{Proceedings of the IEEE/CVF Conference on Computer Vision
  and Pattern Recognition}, 25399--25408.

\bibitem[{Zhu et~al.(2022)Zhu, Zhu, Chu, Zhang, Ji, Wang, and
  Tai}]{zhu2022blind}
Zhu, F.; Zhu, J.; Chu, W.; Zhang, X.; Ji, X.; Wang, C.; and Tai, Y. 2022.
\newblock Blind face restoration via integrating face shape and generative
  priors.
\newblock In \emph{Proceedings of the IEEE/CVF conference on computer vision
  and pattern recognition}, 7662--7671.

\end{thebibliography}

\end{document}